\pdfoutput=1

\documentclass[11pt]{article}

\usepackage[final]{acl}

\usepackage{times}
\usepackage{latexsym}
\usepackage{scalerel,xparse}
\NewDocumentCommand\emojiculture{}{
    \scalerel*{
        \includegraphics{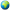}
    }{X}
}
\NewDocumentCommand\emojidefault{}{
    \scalerel*{
        \includegraphics{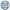}
    }{X}
}

\usepackage{siunitx}
\usepackage{colortbl}
\colorlet{tableheadcolor}{gray!75}
\colorlet{tablerowcolor}{gray!10}
\newcommand{\rowcol}{\rowcolor{tablerowcolor}} %
\usepackage[T1]{fontenc}

\usepackage[utf8]{inputenc}

\usepackage{microtype}

\usepackage{inconsolata}

\usepackage{graphicx}
\usepackage{booktabs}
\usepackage{multirow}
\usepackage{hyperref}
\usepackage{url}
\usepackage{subcaption}
\usepackage{array} 
\usepackage{tcolorbox}
\usepackage{alphabeta}
\usepackage{fancyvrb}
\usepackage{fvextra}

\usepackage[colorinlistoftodos,prependcaption,textsize=tiny]{todonotes}

\title{Vision-Language Models under Cultural and Inclusive Considerations}

\author{
 \textbf{Antonia Karamolegkou\textsuperscript{}},
 \textbf{Phillip Rust\textsuperscript{}},
 \textbf{Yong Cao\textsuperscript{}},
 \\
 \textbf{Ruixiang Cui\textsuperscript{}},
 \textbf{Anders Søgaard\textsuperscript{}},
 \textbf{Daniel Hershcovich\textsuperscript{}}
\\
\\
 \textsuperscript{}Department of Computer Science, University of Copenhagen
\\
 \small{
   \textbf{Correspondence:} \href{mailto:email@domain}{antka@di.ku.dk}
 }
}

\begin{document}
\maketitle
\begin{abstract}

Large vision-language models (VLMs) can assist visually impaired people by describing images from their daily lives. Current evaluation datasets may not reflect diverse cultural user backgrounds or the situational context of this use case. To address this problem, we create a survey to determine caption preferences and propose a culture-centric evaluation benchmark by filtering VizWiz, an existing dataset with images taken by people who are blind. We then evaluate several VLMs, investigating their reliability as visual assistants in a culturally diverse setting. While our results for state-of-the-art models are promising, we identify challenges such as hallucination and misalignment of automatic evaluation metrics with human judgment. We make our survey, data, code, and model outputs publicly available.

\vspace{.3em}
\includegraphics[width=1.25em,height=1.25em]{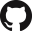}{\hspace{.75em}\parbox{\dimexpr\linewidth-2\fboxsep-2\fboxrule}{\vspace{-5pt} \href{https://github.com/coastalcph/vizwiz-culture}{\texttt{coastalcph/vizwiz-culture}}}}
\end{abstract}

\section{Introduction}

With the increasing integration of AI applications into our lives, it is important to consider human-centered use cases when evaluating such systems. Large multimodal language models are now used as visual assistants for blind and visually impaired individuals. Given that people across different cultures use such applications, it is essential to ensure not only their accuracy and faithfulness \citep{vis_chall, gonzalez2024investigating} but also their cultural representation and inclusion \citep{hershcovich-etal-2022-challenges, shi2024culturebank}.

\begin{figure}[ht]
    \centering
    \includegraphics[width=\linewidth]{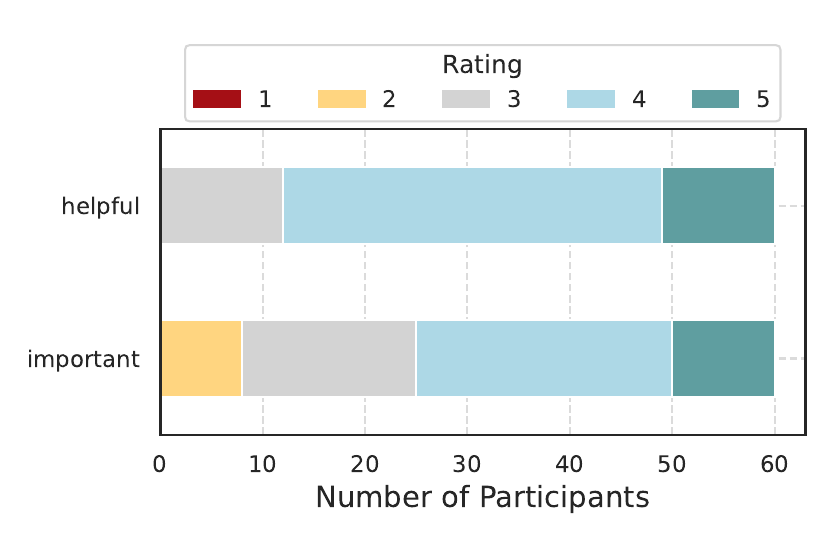}
    \caption{Survey results from people with visual impairments rating \textit{importance} and \textit{helpfulness} of cultural information in image captions. We use a Likert scale from 1 (not important/helpful) to 5 (very important/helpful).}
    \label{fig:viz_survey}
\end{figure}

Existing evaluation benchmarks for VLMs focus primarily on English with few, implicit mutlicultural references. Although multicultural evaluation datasets like MaRVL \citep{liu-etal-2021-visually} and XM3600 \citep{thapliyal-etal-2022-crossmodal} include culture-specific images (e.g., traditional wedding costumes), they also contain images with minimal cultural significance (e.g., a bag of carrots). Consequently, these datasets may not accurately measure the cultural knowledge of VLMs, despite being useful for assessing their multilingual capabilities. Additionally, evaluating these systems as visual assistants presents further challenges due to varying photo quality, user goals, and photo content \citep{Chiu2020AssessingIQ, personal_activities}. Recently, \citet{gonzalez2024investigating} conducted a diary study with blind and low-vision individuals using an AI-powered scene description application, revealing that significant improvements are still needed for satisfying and trustworthy user experiences.

To address both cultural and visual challenges, we first surveyed visually impaired individuals to gather their caption preferences and determine if cultural details are necessary. Then, we filtered an existing dataset with images taken from people who are blind, identifying implicit cultural concepts. This is used as a challenging benchmark to evaluate image captioning performance on cultural images of state-of-the-art models across different prompt settings. With these experiments, we investigate how AI applications, such as image captioning, can foster a more inclusive and culture-aware experience for all.

\paragraph{Background}

Current models are trained without consideration for the subjective perspectives and cultural influences of those who provided the image descriptions \citep{ye2023cultural}. This raises the need for carefully curated sources of data and annotation paradigms that are more culturally aware and inclusive \citep{arora-etal-2023-probing,cao-etal-2023-assessing}. Lately, there has been a growing body of work releasing multicultural multimodal datasets for visiolinguistic reasoning \citep{liu-etal-2021-visually}, text to image generation \citep{liu2023equitable, ventura2023navigating}, and image captioning \citep{thapliyal-etal-2022-crossmodal}.
Beyond the focus on the multilingualism of the captions, concurrent work also addresses the cultural concepts depicted in the images \cite{cao2024exploring,burdalassen2024culturallyawarevisionlanguagemodels,romero2024cvqaculturallydiversemultilingualvisual,mukherjee2024crossroads,bhatia2024localconceptsuniversalsevaluating}. However, they still do not take into account specific use cases, such as visual assistance. \citet{gurari_captioning} released the first image-captioning dataset with photos from people who are blind, and a series of challenges for multimodal systems across different tasks \citep{viz_grand_chall}. After this initiative, there have been many works trying to improve current models for a specific use-case, to assist people with visual disabilities \citep{lessons_learned, ahsan-etal-2021-multi, delloul2023real}. There has also been research in human-computer interaction (HCI) and accessibility on designing image descriptions for visually impaired individuals, primarily focusing on screen readers and functional descriptions of online, publicly available images \citep{Morris2018RichRO, vis_content, screen_readers}. Despite these efforts, there still seems to be a lack of focus on image captioning for the visually impaired \citep{survey_vl}, especially in multi-cultural settings.

\section{Methodology}

We first created a survey seeking to understand the preferences of visually impaired individuals for image captions, focusing on the inclusion of cultural information and the desired level of detail (see Appendix~\ref{app:caption}). We aggregate the participants' assessments of the helpfulness and importance of cultural information in Figure~\ref{fig:viz_survey}. 

We then focused on two lines of contribution: \textbf{(1)} We filtered the VizWiz dataset for implicit cultural concepts. VizWiz is a widely used visual question answering and image captioning dataset representing a real-world use case, where examples consist of images and questions submitted by people who are blind, together with crowdsourced answers and image captions \citep{gurari_captioning}.  The selection of this dataset serves two main purposes. Firstly, it is a challenging dataset specifically tailored to real-world challenges faced by people seeking to access visual information. Secondly, VizWiz might contain implicit cultural references that are currently not captured due to the lack of culture-specific captions.
\textbf{(2)} We evaluated the image captioning performance of state-of-the-art close-sourced and open-sourced models in a culturally diverse setting using our filtered VizWiz dataset. We performed both an automatic scoring of model-generated captions against two sets of annotations using the COCO evaluation package\footnote{\url{https://github.com/tylin/coco-caption}} and a human evaluation.

\subsection{Data Filtering}
To filter the data we hired a total of 165 annotators through the Prolific platform.\footnote{\url{https://www.prolific.com/}} We first asked participants to specify their country of origin, location, and their cultural background. Then, we asked them to retrieve images from the VizWiz dataset visualizer\footnote{\url{https://vizwiz.cs.colorado.edu/VizWiz_visualization/view_dataset.php}} related to their cultural background, provide the image name, the reason they think the image is culture-related, and their preferred caption from the dataset (VizWiz provides five different image captions per image). We also gave them the option to suggest a better caption that includes cultural aspects. After collecting all the culture-specific candidate images, we proceeded to a second step of verification. In this step, we retained only those images that had received consensus agreement from at least two individuals. We collected a total of $324$ images and $648$ captions spanning 60 different identified cultures. It should also be noted that more than 96\% of the annotators suggested a cultural revision of the original captions. We refer to Appendix~\ref{app:human_annot} for further information about the annotation guidelines and data filtering approach and results.

\begin{table*}[ht!]
\centering
\fontsize{14}{18}\selectfont
\resizebox{\textwidth}{!}{%
\begin{tabular}{@{}lllS[table-format=2.1]S[table-format=2.1]@{\footnotesize\,\( \pm \)\,}lS[table-format=2.1]llcS[table-format=2.1]cS[table-format=2.1]llcS[table-format=2.1]cS[table-format=2.1]llcS[table-format=2.1]cS[table-format=2.1]@{}}
\toprule
\textbf{Model}           &  & \multicolumn{5}{c}{\textbf{BLEU-4}}                                                                                                                                            &  &  & \multicolumn{4}{c}{\textbf{METEOR}}                                                                                                                               &  &  & \multicolumn{4}{c}{\textbf{CIDEr}}                                                                                                                                &  &  & \multicolumn{4}{c}{\textbf{SPICE}}                                                                                                                                \\ \midrule
\rowcol\multicolumn{1}{l}{\textit{Prompt}}           & & \multicolumn{2}{c}{\textit{Default}}                                            & \multicolumn{3}{c}{\textit{Cultural}}                                                        &  &  & \multicolumn{2}{c}{\textit{Default}}                                            & \multicolumn{2}{c}{\textit{Cultural}}                                           &  &  & \multicolumn{2}{c}{\textit{Default}}                                            & \multicolumn{2}{c}{\textit{Cultural}}                                           &  &  & \multicolumn{2}{c}{\textit{Default}}                                            & \multicolumn{2}{c}{\textit{Cultural}}                                           \\
\rowcol\multicolumn{1}{l}{\textit{Annotation}}   & & \multicolumn{1}{c}{\normalsize{\textit{Original}}}                 & \multicolumn{1}{c}{\normalsize {\textit{Cultural}}} & \multicolumn{2}{c}{\normalsize{\textit{Original}}}                              & \multicolumn{1}{c}{\normalsize {\textit{Cultural}}} &  &  & \multicolumn{1}{c}{\normalsize{\textit{Original}}}                 & \multicolumn{1}{c}{\normalsize {\textit{Cultural}}} & \multicolumn{1}{c}{\normalsize{\textit{Original}}}                 & \multicolumn{1}{c}{\normalsize {\textit{Cultural}}} &  &  & \multicolumn{1}{c}{\normalsize{\textit{Original}}}                 & \multicolumn{1}{c}{\normalsize {\textit{Cultural}}} & \multicolumn{1}{c}{\normalsize{\textit{Original}}}                 & \multicolumn{1}{c}{\normalsize {\textit{Cultural}}} &  &  & \multicolumn{1}{c}{\normalsize{\textit{Original}}}                 & \multicolumn{1}{c}{\normalsize {\textit{Cultural}}} & \multicolumn{1}{c}{\normalsize{\textit{Original}}}                 & \multicolumn{1}{c}{\normalsize {\textit{Cultural}}}\\ \midrule
BLIP-2          &  & \hspace{0.25em} \underline{8.0\footnotesize{$\pm$ 0.4}}  & 4.8                            & 7.0                   &\footnotesize{0.4}                  & 4.6                            &  &  & \underline{12.6\footnotesize{$\pm$ 0.2}} & 10.2                           & 12.3\footnotesize{$\pm$ 0.3}                  & 10.3                           &  &  & \underline{51.3\footnotesize{$\pm$ 3.2}} & 39.9                           & 44.0\footnotesize{$\pm$ 3.0}                  & 36.7                           &  &  & \underline{13.8\footnotesize{$\pm$ 0.4}} & 12.5                           & 12.8\footnotesize{$\pm$ 0.5}                  & 11.5                           \\
InstructBLIP  &  & 14.0\footnotesize{$\pm$ 0.5}                  & 8.7                            & \underline{14.1} & \underline{\footnotesize{0.4}} & 9.0                            &  &  & 17.3\footnotesize{$\pm$ 0.3}                  & 13.2                           & \underline{17.7\footnotesize{$\pm$ 0.3}} & 13.3                           &  &  & 77.1\footnotesize{$\pm$ 3.4}                  & 60.0                           & \underline{78.8\footnotesize{$\pm$ 3.2}} & 60.2                           &  &  & \underline{18.5\footnotesize{$\pm$ 0.4}} & 15.6                           & 18.2\footnotesize{$\pm$ 0.5}                  & 14.9                           \\
Idefics2            &  & \underline{12.0\footnotesize{$\pm$ 0.5}} & 10.1                           & 9.8                   &\footnotesize{0.5}                  & 10.7                           &  &  & 18.1\footnotesize{$\pm$ 0.3}                  & 15.1                           & \underline{18.9\footnotesize{$\pm$ 0.3}} & 17.1                           &  &  & \underline{80.2\footnotesize{$\pm$ 1.9}} & 78.4                           & 74.1\footnotesize{$\pm$ 2.2}                  & 78.2                           &  &  & 18.0\footnotesize{$\pm$ 0.5}                  & 16.7                           & \underline{18.8\footnotesize{$\pm$ 0.2}} & 17.8                           \\
LLaVA-1.6   &  & 10.0\footnotesize{$\pm$ 0.5}                  & \underline{11.4}          & 6.7                   &\footnotesize{0.3}                  & 7.7                            &  &  & \underline{18.9\footnotesize{$\pm$ 0.4}} & 17.3                           & 18.4\footnotesize{$\pm$ 0.3}                  & 17.0                           &  &  & 60.2\footnotesize{$\pm$ 2.3}                  & \underline{75.2}          & 40.3\footnotesize{$\pm$ 1.7}                  & 56.3                           &  &  & 16.3\footnotesize{$\pm$ 0.6}                  & \underline{16.5}          & 15.8\footnotesize{$\pm$ 0.5}                  & 15.4                           \\
Gemini-1.5-Pro          &  & 10.8\footnotesize{$\pm$ 0.3}                  & \underline{14.1}          & 5.8                   &\footnotesize{0.1}                  & 8.7                            &  &  & 20.8\footnotesize{$\pm$ 0.4}                  & \underline{21.3}          & 18.2\footnotesize{$\pm$ 0.1}                  & 21.0                           &  &  & 71.5\footnotesize{$\pm$ 2.1}                  & \underline{88.8}          & 14.8\footnotesize{$\pm$ 0.5}                  & 34.1                           &  &  & 19.6\footnotesize{$\pm$ 0.4}                  & \underline{21.6}          & 14.9\footnotesize{$\pm$ 0.3}                  & 17.7                           \\
GPT-4o                  &  & 11.9\footnotesize{$\pm$ 0.6}                  & \textbf{\underline{16.4}} & 8.1                   &\footnotesize{0.3}                  & 12.2                           &  &  & 22.4\footnotesize{$\pm$ 0.4}                  & \textbf{\underline{23.4}} & 19.9\footnotesize{$\pm$ 0.3}                  & 22.6                           &  &  & 66.8\footnotesize{$\pm$ 2.8}                  & \textbf{\underline{99.8}} & 40.4\footnotesize{$\pm$ 1.0}                  & 72.8                           &  &  & 19.1\footnotesize{$\pm$ 0.4}                  & \textbf{\underline{21.8}} & 16.6\footnotesize{$\pm$ 0.3}                  & 20.1                           \\ \bottomrule
\end{tabular}%
}
\caption{Performance of various VLMs on our filtered VizWiz dataset across captioning prompts (default \& culture-specific) and annotations (original \& culture-specific). We use 2 reference annotations per image. Since the original VizWiz has 5 annotations per image, we report the mean and standard deviation over all 10 combinations with two references. We \underline{underline} the best result for each model and display the top result for each metric in \textbf{bold}.
}
\label{tab:coco_eval}
\end{table*}

\subsection{Models and evaluation}
We conducted experiments on the image captioning task in the zero-shot setting, in which a pretrained model is queried to produce a textual description for an image without finetuning on the same dataset. We relied on four commonly used open-access models:\footnote{We used implementations and model weights from HuggingFace \citep{wolf-etal-2020-transformers}.} BLIP-2 6.7B \citep{li2023blip2} with OPT as LLM backbone \citep{zhang2022opt}, InstructBLIP 7B \citep{dai2023instructblip} with Vicuna backbone \citep{vicuna2023}, Idefics2 8B \citep{idefics2}, and LLaVa-1.6 7B \citep{llava} with Mistral backbone \citep{jiang2023mistral7b}.
We also used two state-of-the-art closed-access models: GPT-4o \citep{gpt4o} and Gemini Pro 1.5 \citep{team2024gemini}.
For all of these models, we experimented with two different prompt types including a culture-specific prompt following \citet{shi2024culturebank} and a default captioning prompt taken from \citet{dai2023instructblip}. The exact prompts can be found in App~\ref{app:prompt}. 
We evaluated the model-generated captions in two ways: (1) via the COCO evaluation suite and (2) through human evaluation. The COCO evaluation suite was first introduced by \citep{chen2015microsoft} as a framework to assess image captions using numerous automatic metrics, including BLEU \citep{papineni-etal-2002-bleu}, CIDEr \citep{cider}, METEOR \citep{denkowski-lavie-2014-meteor}, and SPICE \citep{DBLP:conf/eccv/AndersonFJG16}. For consistency with our culture-specific re-annotations (two captions per image), we also used two reference captions per image to score models on the original annotations. Since each image has five original captions, we report aggregate results over all ten two-caption combinations. Our human evaluation had two stages. In the first stage, we asked 60 participants to determine if a caption is accurate (on a binary scale) given the corresponding image. In the second stage, we asked the same participants to rank all captions (human-generated, and model-generated) according to their preference. We did not make the annotators aware that one caption was model-generated to minimize bias. We provide further details on the human evaluation in Appendix~\ref{app:human_eval}.

\section{Results}

\paragraph{Automatic evaluation}

We present the results of our automatic evaluation of model-generated captions in
Table~\ref{tab:coco_eval}. Note that due to using two reference captions per image, results for the original annotations are slightly different than when using all five at once; we report the latter in Appendix~\ref{app:full_orig_results} for completeness.

\begin{figure*}
    \centering
    \includegraphics[width=\textwidth]{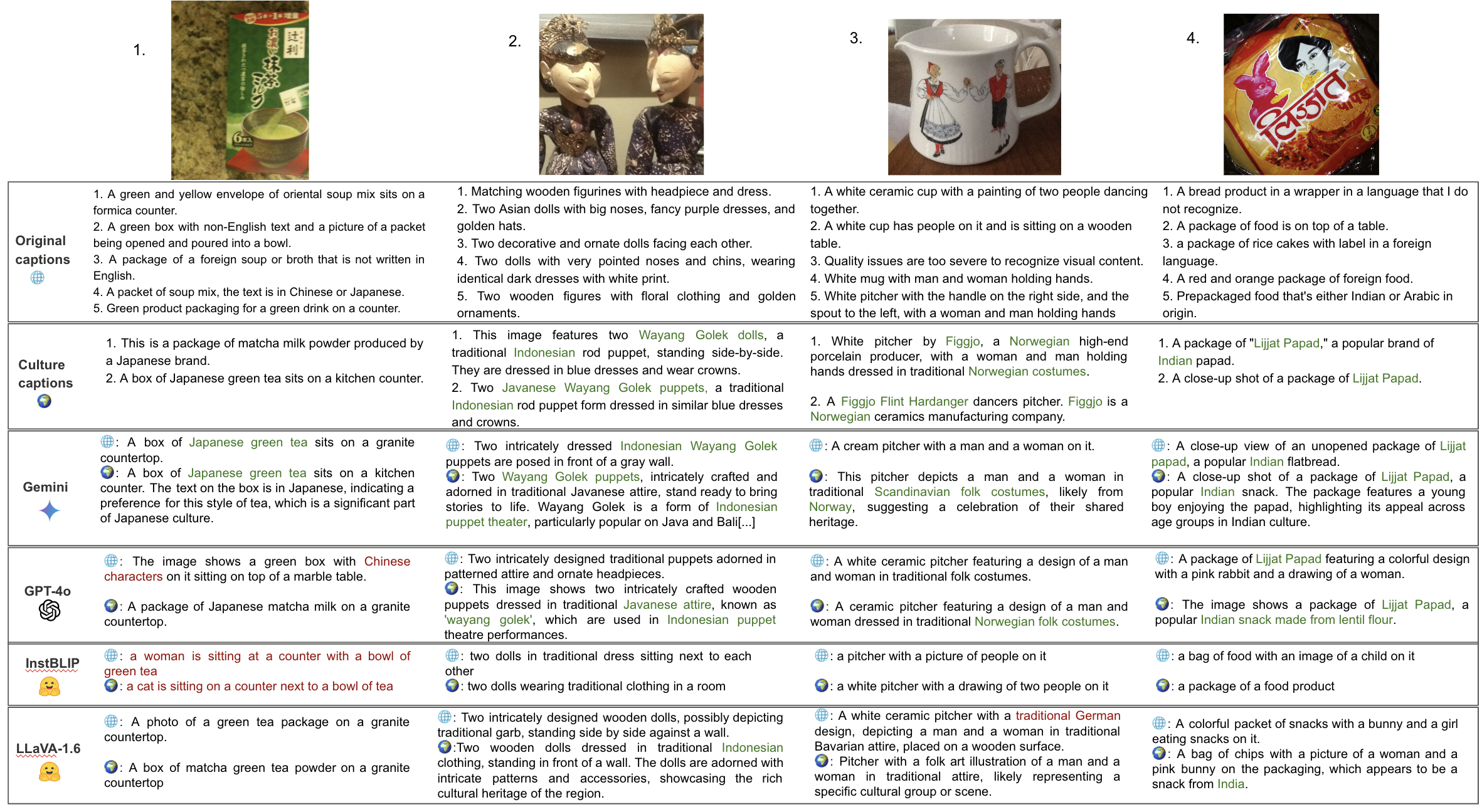}
    \caption{Examples of various images from the filtered VizWiz dataset with the original (\emojidefault) and culture-specific (\emojiculture) annotations, and generated captions from Gemini-1.5-Pro, GPT-4o, InstructBLIP, and LLaVA-1.6 with default (\emojidefault) and culture-specific (\emojiculture) prompting.}
    \label{fig:multivizexamples}
\end{figure*}

\begin{figure}[ht]
    \centering
    \includegraphics[width=0.95\linewidth]{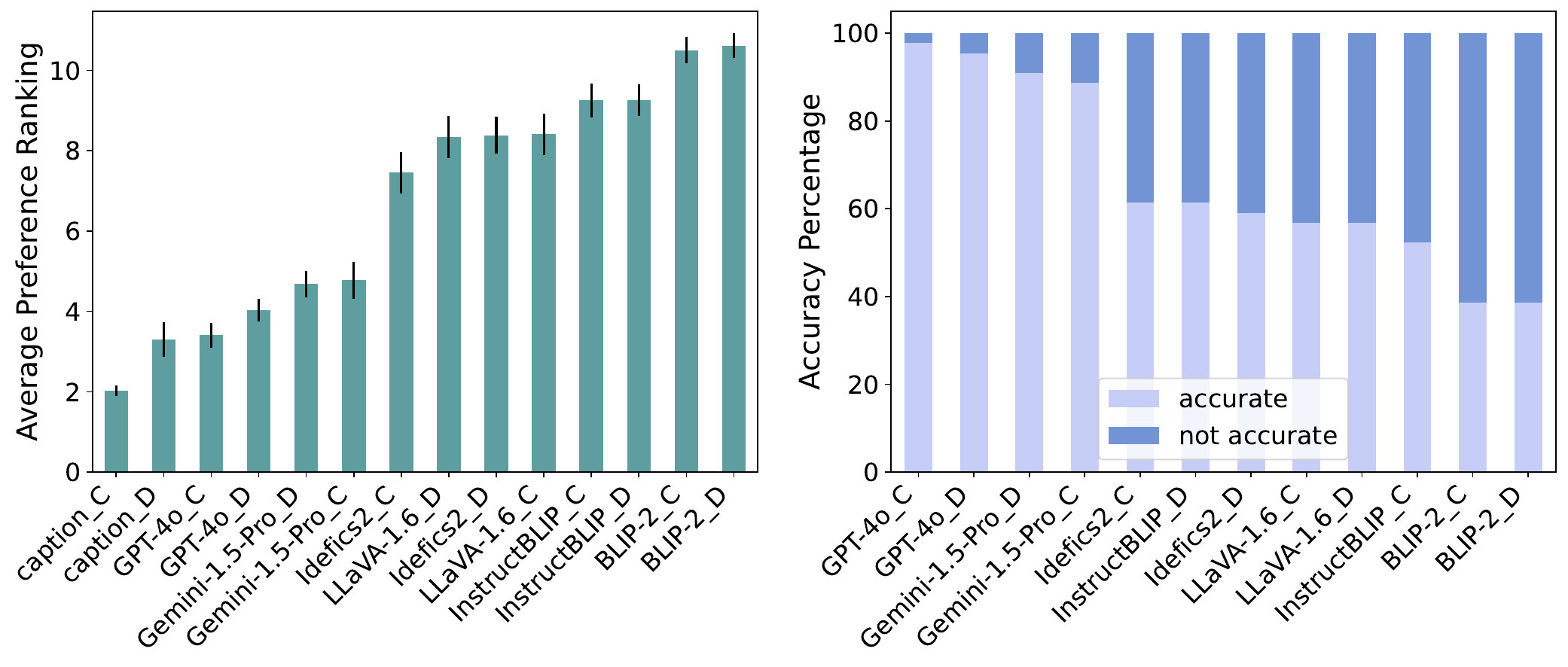}
    \caption{Results of the human evaluation for 100 images and their captions selected at random from the filtered VizWiz dataset. The left plot shows the preference score (participants were asked to rank the captions; lower is better). The right plot shows the accuracy evaluation (participants were asked to assess whether a caption is accurate; higher is better). \emph{`\_D'} and \emph{`\_C'} denote default and culture-specific prompting, respectively.}
    \label{fig:viz_eval}
\end{figure}

As expected, the closed-access models (Gemini and GPT-4o) score best overall. Slightly lower performance is achieved by the instruction-tuned open-access VLMs (LLaVa, Idefics2, and InstructBLIP). BLIP-2, which has not been instruction-tuned, is lagging behind across all metrics. Since VizWiz is naturally noisy due to the high ratio of low-quality, blurry images, the increased scale and overall multimodal reasoning capabilities of the closed-source models appear to give a significant advantage.

Strikingly, Gemini and GPT-4o achieve much better performance on our newly annotated captions that include cultural information than on the original captions (e.g., $11.9$ vs. $16.4$ BLEU-4 and $66.8$ vs. $99.8$ CIDEr for GPT-4o with the default prompt), while we observe the opposite for the open-access models (e.g, $14.0$ vs. $8.7$ BLEU-4 and $77.1$ vs. $60.0$ CIDEr for InstructBLIP with the default prompt). One possible explanation is that the closed-source models have been tuned to generate more descriptive captions that are aligned better with human preferences and our cultural caption annotations, whereas the open-access models have been tuned to generate slightly more concise captions that align well with benchmark datasets like COCO Captions. Our new cultural annotations are also guaranteed to not have leaked into the VLMs' training data, thus favoring more objectively capable models such as GPT-4o.

Next, while individual models (Idefics2 and InstructBLIP in particular) seem amenable to cultural prompting, leading to improved performance even on the original image captions, the cultural prompting strategy is overall largely ineffective at improving performance on the cultural captions.
This result may be due to the models' tendency for sycophantic behavior and them being primed to point out cultural information over other relevant content in the image \citep{sharma2023understandingsycophancylanguagemodels}. Alternatively, cultural prompting might elicit more verbose captions that are disfavored by the automatic evaluation metrics, in which case the automatic evaluation results paint an incomplete and potentially misleading picture.

\paragraph{Human evaluation}

The results of the human evaluation are shown in Figure~\ref{fig:viz_eval}. In line with the automatic metrics, our human annotators tend to prefer the captions produced by closed-access models, GPT-4o and Gemini-Pro, with the BLIP-family models having the lowest ranking. The former are rated as accurate in more than 90\% of the cases, while the latter are deemed inaccurate in more than half of the cases. Despite the strong performance of the closed-access models, our preference comparison also shows that the culture-specific human-annotated captions are still preferred over all of the models, suggesting there is ample room for improvement.

In spite of the often stark differences in automatic evaluation scores between cultural and default prompting (with a preference for the latter), human participants prefer the model generations obtained via cultural prompting in 4/6 cases (for both the ranking and the accuracy assessment), supporting our hypothesis that cultural prompting simply elicits an answer format that is disfavored by automatic metrics.
 
Overall, our results are promising in regard to the reliability of VLMs at zero-shot generating captions that are accurate and useful to users who are blind in culturally diverse scenarios.

\section{Further Analysis}

To further analyze our results and assess the model-generated captions in a more fine-grained manner, we manually inspected all generated captions for our $324$ images filtered VizWiz dataset and provided some examples in Figure \ref{fig:multivizexamples}. 

We find that InstructBLIP and BLIP-2 captions tend to be very short, lack a lot of information, and are often irrelevant hallucinations. This is, to an extent, expected as we perform zero-shot captioning, so the models are not necessarily accustomed to the desired captioning style. In this case, few-shot prompting or finetuning the models would likely improve model performance \citep{NEURIPS2020_1457c0d6, manas-etal-2023-mapl, rita}. The closed-access models, in contrast, largely provide further or more useful and culture-specific details about the image than given by the human captioners. They also seem to provide more accurate captions compared to the open-sourced models. These points may explain why GPT-4o and Gemini-1.5-Pro and were overall preferred in our human evaluation. 

Overall, we observed that the closed-access models can transcribe various language scripts from books, food or beverage packages, giving them an advantage over the smaller models. In most cases, in both culture-specific and default prompts, the models can identify culture-specific beverages like Japanese matcha tea, Chinese jelly grass or lychee juice, and food such as the Indian lijjat papad, Japanese mochi, Tom Kha Gai Thai soup, Korean kimchi, etc. There are also cases where they identify religious or folk items like the Wayang Golek puppets, a jar with traditional Norwegian costumes, or a delft plaque with traditional Dutch costumes. 

There is, however, a tendency to generate longer text in the culture-specific prompts by adding generic phrases such as `\textit{hinting at the drink's cultural origin}', `\textit{suggesting a celebration of their shared heritage}', `\textit{highlighting its appeal across age groups in Indian culture}', etc.
The most challenging cases for the closed-source models seem to be foreign currencies (especially the Arabic ones), historic figures, and paintings. For example, models seem to confuse Bahraini, Jordan, and Egyptian banknotes, and they do not recognize the Chinese historical figure of Sun Yat-sen, or paintings of Joan Miró or Frederick Morgan. We provide further examples in Appendix~\ref{app:errors}.

\section{Discussion}

Given the current integration of VLMs as virtual assistants for people who seek sighted support, their performance on culture-specific image captioning seems promising. Examples from our error analysis and case studies highlight some remaining challenges. Measured by automatic evaluation metrics, the performance of the models is overall relatively low compared to results in existing studies evaluating (finetuned) VLMs for image captioning on the full VizWiz and other datasets \citep{gurari_captioning, chen2023pali, wang2022git}. On the other hand, our human evaluation and error analysis show that the generated captions by Gemini-1.5-Pro and GPT-4o are accurate and preferred in many cases. There also seems to be an extended hallucination problem, which remains an existing major challenge not only for VLMs \citep{li-etal-2023-evaluating} but across various language model applications \citep{bang-etal-2023-multitask, hallucin}.

\section{Conclusion}

We evaluated the cultural performance of various models on image captioning using a multicultural dataset tailored to a real-world use case. Although the performance of state-of-the-art closed-source models is promising, there is plenty room for improvement. Examples from our error analysis provide insights into the models' performance, helping us identify some of their weak spots. In our use case, we find that automatic evaluation metrics might not be fully representative of model performance, and therefore encourage researchers to reconsider a more comprehensive assessment framework. For future work, we aim to extend our small filtered cultural dataset by including question-answering tasks with POV cultural questions. 

\section*{Limitations}

Our work focuses primarily on data curation and empirical analysis of large multimodal language models. Our survey, while aimed at determining caption preferences, may not capture the full range of needs and preferences of all people with visual impairment. 
Further, through our analysis, we gained insights into some weak spots with respect to what cultures and cultural concepts are well recognized by the models. However, since we use a finite amount of data, there might be a data bias in identifying particular cultures or cultural concepts as problematic. Lastly, cultural complexities and variations make it difficult to develop a standardized approach to cultural inclusion in AI. We do, however, hope that our culture-centric approach in the data filtering and annotation process can serve as an initial step towards evaluating and understanding the cultural awareness and abilities of vision-language models for real-world uses. 

\section*{Ethics Statement}

The motivation behind this study is that large vision-language models have rapidly become mainstream and are used even by those who seek sighted support and cannot easily assess model hallucinations or inaccuracies. The primary purpose of our experiments is to assess the performance of vision-language models in the task of image captioning using a multicultural dataset of images taken from people who are blind. However, it is crucial to recognize that results from our current filtered dataset may not be representative of model performance across cultures. Furthermore, our refined dataset might retain biases present in the original source dataset.

We find it improbable that our experiments and the filtered dataset will meaningfully benefit those intending to create deceptive models for malicious purposes. Additionally, the VizWiz dataset may lack coverage of highly specific subjects, offering only a general overview of factual topics. People who intend to use our resources, however, should state their purpose of usage and be accountable for their own work.

\section*{Acknowledgments}

\bibliography{acl_latex}

\newpage

\appendix

\section{Survey on Caption Preferences}
\label{app:caption}

We created a survey aiming to understand the preferences of individuals who seek sighted support regarding image captioning. Our interest was particularly focused on whether they prefer image captions to include cultural information, and how detailed they prefer the descriptions to be. We published our survey through the Prolific platform, by choosing  60 participants with an equal gender sample of and representative across countries compensated with \$18 per hour. We also added a screener and selected participants without corrected/normal vision. Overall, the participants were positive regarding the helpfulness and importance of cultural information in the captions with average ratings of 4.1 and 3.9, respectively.\footnote{The scale is from 1. Not important/helpful at all to 5. Very important/helpful)} Participants also tended to prefer short captions compared to longer ones.

\section{VizWiz Data Filtering -- Human Annotation}
\label{app:human_annot}

As mentioned in the experimental set-up section, to filter the data we created a survey through the Prolific annotation platform. All annotators were compensated with 18\$ per hour. We ran this survey 4 times asking for 40 participants each time. 

\begin{figure}[ht]
    \centering
    \includegraphics[width=0.8\linewidth]{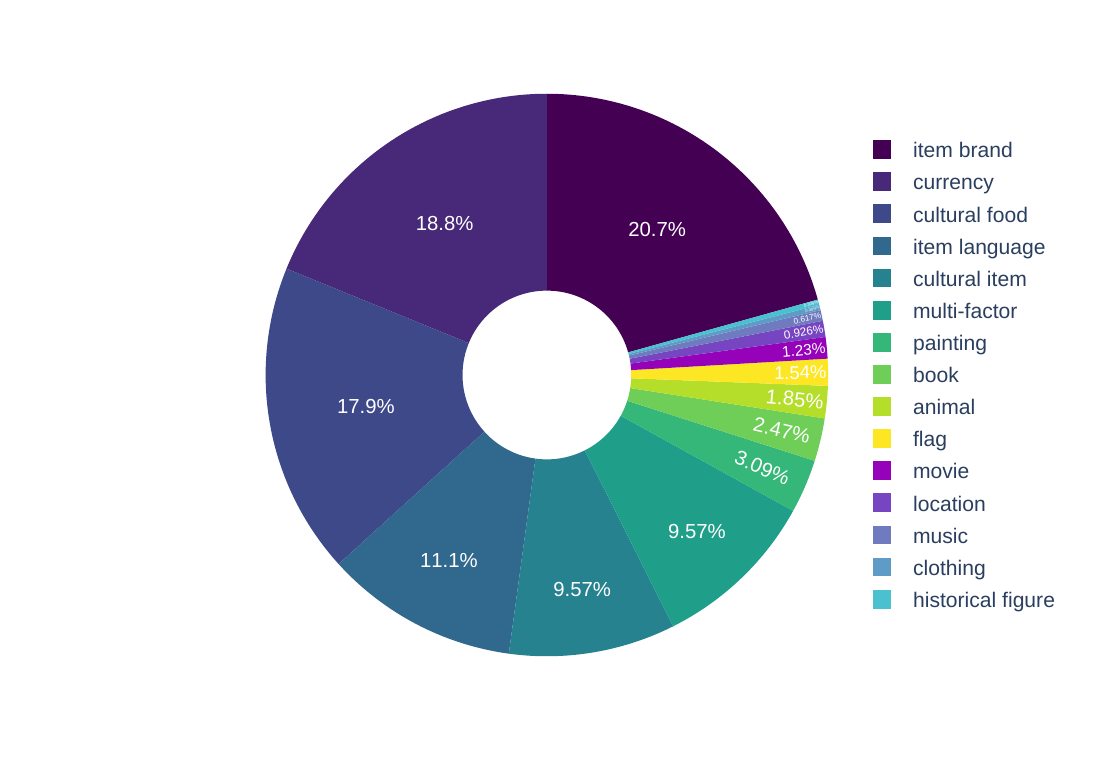}
    \caption{Distribution of the factors/indicators that lead the annotators to select a specific image as culture-related and specify the corresponding culture.}
    \label{fig:factors}
\end{figure}

We asked people to identify images from the VizWiz dataset based on their cultural background, provide an original and a corrected caption, and specify the reason they selected the image as culture-specific. We grouped the reasons that the annotators provided for selecting culture-specific images in Figure~\ref{fig:factors}. 

\begin{figure}[ht]
    \centering
    \includegraphics[width=\linewidth]{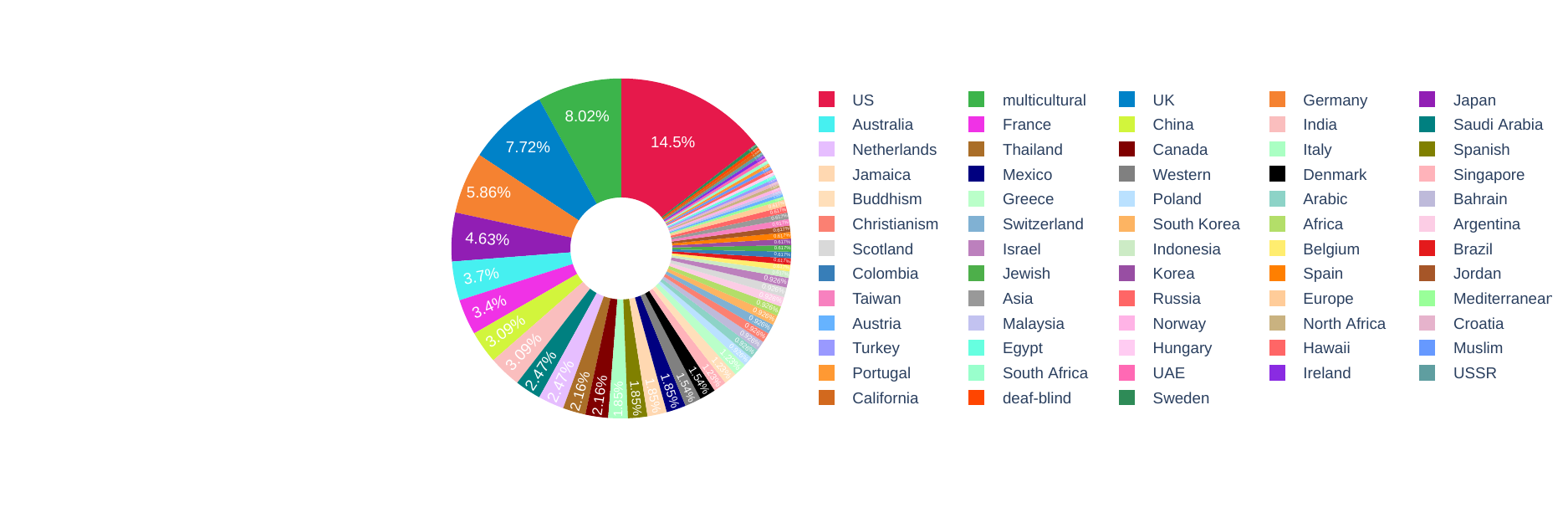}
    \caption{Distribution of the cultural concepts identified in the VizWiz dataset by the annotators.}
    \label{fig:concepts}
\end{figure}

The cultural concepts identified by our annotators can be found in Figure~\ref{fig:concepts}. 

The full annotation guidelines were the following:

\begin{tcolorbox}[colback=white,colframe=blue,sharp corners]

Creating datasets that reflect a variety of cultures is a challenging task. This is why we will try to filter an existing dataset. Your task is to find culture-related images from a dataset called VizWiz. You need to:

- Visit the dataset website[link]. 
- Browse the dataset or use the search bars on the left side of the page and search key-terms related to your culture 'Within visual question', 'Within visual answer' or 'Within captions'.
- Try to find an image that is related to your culture/cultural background (i.e. food brand, currency, books, culture-specific locations etc.)
- Provide your answers to the 5 following questions.

\begin{enumerate}
    \item  Copy and paste the image name (VizWiz\_train\_**number**.jpg).

    \item Based on your cultural background, specify what culture you think is the image related to.

    \item Select a caption for the image from the suggested Image Captions.

    \item Do you have a better suggestion for the image caption? To guide your caption generation, imagine that you are
describing the image to a visually impaired friend. The caption
should explain the whole image, including all the main objects,
activities, and their relationships, and reflect the culture information of the image.

    \item Provide a reason as to why the image is culture-specific.
\end{enumerate}

\end{tcolorbox}

\begin{figure}[!ht]
    \centering
    \begin{subfigure}{0.45\textwidth}
        \includegraphics[width=\linewidth]{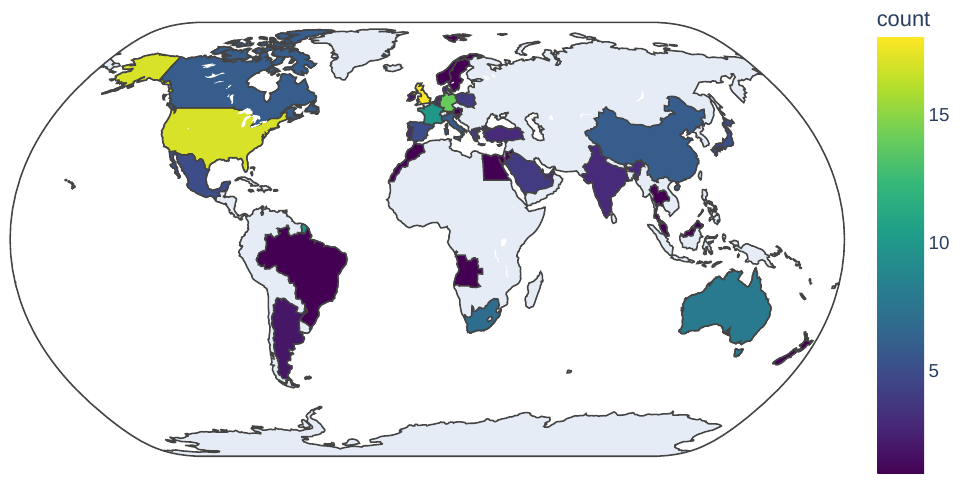}
        \caption{Distribution of the current country location of the annotators.}
        \label{fig:location}
    \end{subfigure}
    \hfill
    \begin{subfigure}{0.45\textwidth}
        \includegraphics[width=\linewidth]{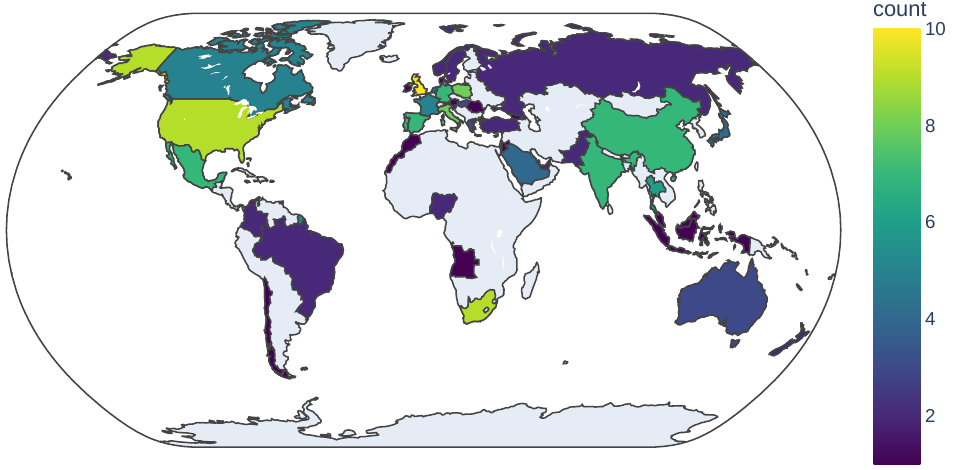}
        \caption{Distribution of the country of origin of the annotators.}
        \label{fig:origin}
    \end{subfigure}
    \caption{Plots as subfigures.}
    \label{fig:subfigures}
\end{figure}

After this, we collected information about the annotators' cultural backgrounds. We asked for both home-country of origin and current country location information since sometimes both can affect our cultural beliefs and practices. The distribution of the annotators counties of origin and location are presented in Figure~\ref{fig:origin}.

\begin{tcolorbox}[colback=white,colframe=blue,sharp corners]
The last step is to answer some final questions about your cultural background, and age. We do not collect any other personal information. Your answers will only be used for statistical research purposes. 

\begin{itemize}
    \item What is your country of origin that you consider your 'home', influencing your cultural beliefs and other aspects of your identity?
    \item Is there a country in which you are currently located for a long period of time?
    \item How old are you? Fill in years in numbers.
\end{itemize}
\end{tcolorbox}

After collecting all the responses, we kept only the images where at least two annotators agreed to select the image as culture-specific. After this extra validation, we resulted in a total of $324$ images spanning 60 different identified cultures. We compared the similarity between the suggested captions by the annotators and the original VizWiz captions and the results can be found in Table~\ref{tab:cap_comp} indicating a high similarity between the new culture-specific suggested captions.

\begin{table}[ht]
\centering
\fontsize{14}{18}\selectfont
\sisetup{table-format = 3.2}
\resizebox{0.7\columnwidth}{!}{%
\begin{tabular}{@{}cccc@{}}
\toprule
Captions             & BLEU-4   & ROUGE-L & F1  \\ \midrule
Culture-specific  & 37.10  & 61.90  & 93.0 \\ \bottomrule
\end{tabular}%
}
\caption{Results from comparing the culture-specific captions of the two annotators against the five original VizWiz captions.}
\label{tab:cap_comp}
\end{table}

\newpage

\section{Model Overview}

We list models with their API identifiers in Table~\ref{tab:api_identifiers} below.

\begin{table}[htb!]
\fontsize{14}{18}\selectfont
\centering
\resizebox{0.48\textwidth}{!}{%
\begin{tabular}{lll}
\toprule
Name           & Identifier                                                                                                  & Reference                   \\ \midrule
BLIP-2         & \href{https://huggingface.co/Salesforce/blip2-opt-6.7b}{\texttt{Salesforce/blip2-opt-6.7b}}                 & \citet{li2023blip2}         \\
InstructBLIP   & \href{https://huggingface.co/Salesforce/instructblip-vicuna-7b}{\texttt{Salesforce/instructblip-vicuna-7b}} & \citet{dai2023instructblip} \\
Idefics2       & \href{https://huggingface.co/HuggingFaceM4/idefics2-8b}{\texttt{HuggingFaceM4/idefics2-8b}}                 & \citet{idefics2}            \\
LLaVA-1.6      & \href{https://huggingface.co/llava-hf/llava-v1.6-mistral-7b-hf}{\texttt{llava-hf/llava-v1.6-mistral-7b-hf}} & \citet{llava}               \\
Gemini-1.5-Pro & \href{https://deepmind.google/technologies/gemini/pro/}{\texttt{gemini-1.5-pro-preview-0514}}               & \citet{team2024gemini}      \\
GPT-4o         & \href{https://openai.com/index/hello-gpt-4o/}{\texttt{gpt-4o-2024-05-13}}                                   & \citet{gpt4o} \\ \bottomrule             
\end{tabular}%
}
\caption{Overview of models used in this study}
\label{tab:api_identifiers}
\end{table}

\section{Model Prompting}
\label{app:prompt}

We provide the templates we used to prompt our models. The default templates have been sourced from \citet{dai2023instructblip} and \citet{shi2024culturebank}.

\begin{table*}[h!]
\centering
\begin{tabular}{@{} m{\linewidth} @{}}
\toprule
\rowcol \small{\textbf{Default prompting}}
\\
\begin{Verbatim}[breaklines=True,commandchars=\\\{\},fontsize=\scriptsize]
<Image> A short image description:
\end{Verbatim}
\\
\rowcol \begin{Verbatim}[breaklines=True,commandchars=\\\{\},fontsize=\scriptsize]
<Image> Write a caption that describes the photo.

Format your response in JSON as follows:
\{
  "caption": "Caption for the image"
\}
\end{Verbatim}
\\
\begin{Verbatim}[breaklines=True,commandchars=\\\{\},fontsize=\scriptsize]
<Image> A photo of
\end{Verbatim}
\\ 
\rowcol \begin{Verbatim}[breaklines=True,commandchars=\\\{\},fontsize=\scriptsize]
<Image> Can you briefly describe the content of the image?
\end{Verbatim}
\\
\begin{Verbatim}[breaklines=True,commandchars=\\\{\},fontsize=\scriptsize]
<Image> Write a caption that describes the photo.
\end{Verbatim}
\\
\toprule
\rowcol \small{\textbf{Culture-specific prompting}}
\\
\begin{Verbatim}[breaklines=True,commandchars=\\\{\},fontsize=\scriptsize]
<Image> A short, culture-aware image description:
\end{Verbatim}
\\
\rowcol \begin{Verbatim}[breaklines=True,commandchars=\\\{\},fontsize=\scriptsize]
<Image> Cultural information encompasses content that showcases the distinctive characteristics, artifacts, or manifestations of a specific group, community, or region. 
This includes, but is not limited to, practices, behaviors, norms, values, beliefs, habits, customs, architectural styles, environmental engagements, and any other elements that are emblematic of a particular cultural setting. 
It does not include generic information or widespread practices that are not distinctly tied to a specific cultural identity. 

For this task, consider information as "cultural" if:
1. It is associated with or characteristic of a specific identified group (e.g., Americans, Italians, midwestern Americans, etc.).
2. It reveals a unique aspect of that group’s way of life, including social conventions, physical creations, or interactions with their surroundings that are not typically seen in other cultures.
3. It provides insight into the cultural uniqueness, whether through social practices, material culture, or other culturally significant elements.

Please exclude generic or ubiquitous statements or observations that do not clearly relate to the unique cultural context of a specific group.

Given this image, do two things:
1. Determine whether the provided example contains cultural information.
2. Write a caption that describes the photo and includes the cultural information extracted.

Format your response in JSON as follows:
\{
  "caption": "Caption for the image",
  "is_cultural": true/false,
  "justification": "Why or why not the image contains cultural information"
\}
\end{Verbatim}
\\
\begin{Verbatim}[breaklines=True,commandchars=\\\{\},fontsize=\scriptsize]
<Image> Write a caption that describes the photo and includes any cultural information present.
\end{Verbatim}
\\
\bottomrule
\end{tabular}
\caption{Image captioning templates used to prompt our models.}
\end{table*}

\section{Vizwiz Results -- 5 Original References}
\label{app:full_orig_results}
We report model performance on our filtered VizWiz dataset when using all five original captions per image (rather than combinations of two references at a time) in Figure~\ref{fig:viz_results_5cap}.

\begin{table*}[ht!]
\centering
\fontsize{14}{18}\selectfont
\resizebox{\textwidth}{!}{%
\begin{tabular}{@{}lS[table-format=2.1]S[table-format=2.1]S[table-format=2.1]S[table-format=2.1]S[table-format=2.1]S[table-format=2.1]S[table-format=2.1]S[table-format=2.1]S[table-format=2.1]S[table-format=2.1]S[table-format=2.1]S[table-format=2.1]S[table-format=2.1]S[table-format=2.1]S[table-format=2.1]S[table-format=2.1]S[table-format=2.1]S[table-format=2.1]S[table-format=2.1]S[table-format=2.1]@{}}
\toprule
\textbf{Model}      &           & \multicolumn{4}{c}{\textbf{BLEU-4}}                                          &           & \multicolumn{4}{c}{\textbf{METEOR}}                                          &           & \multicolumn{4}{c}{\textbf{CIDEr}}                                           &           & \multicolumn{4}{c}{\textbf{SPICE}}                                           \\ \midrule
\rowcol\textit{Prompt}     & \textit{} & \multicolumn{2}{c}{\textit{Default}} & \multicolumn{2}{c}{\textit{Cultural}} & \textit{} & \multicolumn{2}{c}{\textit{Default}} & \multicolumn{2}{c}{\textit{Cultural}} & \textit{} & \multicolumn{2}{c}{\textit{Default}} & \multicolumn{2}{c}{\textit{Cultural}} & \textit{} & \multicolumn{2}{c}{\textit{Default}} & \multicolumn{2}{l}{\textit{Cultural}} \\
\rowcol\textit{Annotation} &           & \multicolumn{4}{c}{\textit{\normalsize{Original-Full}}}                      &           & \multicolumn{4}{c}{\textit{\normalsize{Original-Full}}}                      &           & \multicolumn{4}{c}{\textit{\normalsize{Original-Full}}}                      &           & \multicolumn{4}{c}{\textit{\normalsize{Original-Full}}}                      \\ \midrule
BLIP-2              &           & \multicolumn{2}{S[table-format=2.1]}{\underline{14.9}}             & \multicolumn{2}{S[table-format=2.1]}{12.1}              &           & \multicolumn{2}{S[table-format=2.1]}{\underline{16.1}}             & \multicolumn{2}{S[table-format=2.1]}{15.5}              &           & \multicolumn{2}{S[table-format=2.1]}{\underline{51.7}}             & \multicolumn{2}{S[table-format=2.1]}{44.3}              &           & \multicolumn{2}{S[table-format=2.1]}{\underline{10.6}}             & \multicolumn{2}{S[table-format=2.1]}{9.9}               \\
InstructBLIP        &           & \multicolumn{2}{S[table-format=2.1]}{\textbf{\underline{25.3}}}             & \multicolumn{2}{S[table-format=2.1]}{24.5}              &           & \multicolumn{2}{S[table-format=2.1]}{22.0}             & \multicolumn{2}{S[table-format=2.1]}{\underline{22.1}}              &           & \multicolumn{2}{S[table-format=2.1]}{77.4}             & \multicolumn{2}{S[table-format=2.1]}{\underline{78.9}}              &           & \multicolumn{2}{S[table-format=2.1]}{\underline{15.0}}             & \multicolumn{2}{S[table-format=2.1]}{\underline{15.0}}              \\
Idefics2            &           & \multicolumn{2}{S[table-format=2.1]}{\underline{20.8}}             & \multicolumn{2}{S[table-format=2.1]}{16.7}              &           & \multicolumn{2}{S[table-format=2.1]}{22.2}             & \multicolumn{2}{S[table-format=2.1]}{\underline{23.3}}              &           & \multicolumn{2}{S[table-format=2.1]}{\textbf{\underline{82.0}}}             & \multicolumn{2}{S[table-format=2.1]}{76.1}              &           & \multicolumn{2}{S[table-format=2.1]}{15.1}             & \multicolumn{2}{S[table-format=2.1]}{\underline{16.6}}              \\
LLaVA-1.6           &           & \multicolumn{2}{S[table-format=2.1]}{\underline{17.3}}             & \multicolumn{2}{S[table-format=2.1]}{11.8}              &           & \multicolumn{2}{S[table-format=2.1]}{\underline{23.3}}             & \multicolumn{2}{S[table-format=2.1]}{22.1}              &           & \multicolumn{2}{S[table-format=2.1]}{\underline{60.9}}             & \multicolumn{2}{S[table-format=2.1]}{40.5}              &           & \multicolumn{2}{S[table-format=2.1]}{\underline{15.3}}             & \multicolumn{2}{S[table-format=2.1]}{\underline{15.3}}              \\
Gemini-1.5-Pro      &           & \multicolumn{2}{S[table-format=2.1]}{\underline{18.6}}             & \multicolumn{2}{S[table-format=2.1]}{9.9}               &           & \multicolumn{2}{S[table-format=2.1]}{\underline{25.5}}             & \multicolumn{2}{S[table-format=2.1]}{21.9}              &           & \multicolumn{2}{S[table-format=2.1]}{\underline{73.0}}             & \multicolumn{2}{S[table-format=2.1]}{15.0}              &           & \multicolumn{2}{S[table-format=2.1]}{\underline{18.0}}             & \multicolumn{2}{S[table-format=2.1]}{15.3}              \\
GPT-4o              &           & \multicolumn{2}{S[table-format=2.1]}{\underline{20.5}}             & \multicolumn{2}{S[table-format=2.1]}{13.8}              &           & \multicolumn{2}{S[table-format=2.1]}{\textbf{\underline{27.4}}}             & \multicolumn{2}{S[table-format=2.1]}{24.4}              &           & \multicolumn{2}{S[table-format=2.1]}{\underline{67.7}}             & \multicolumn{2}{S[table-format=2.1]}{41.0}              &           & \multicolumn{2}{S[table-format=2.1]}{\textbf{\underline{18.4}}}             & \multicolumn{2}{S[table-format=2.1]}{16.6}              \\ \bottomrule
\end{tabular}%
}
\caption{Performance of various VLMs on our filtered VizWiz dataset across captioning prompts (default \& culture-specific) using the five original reference captions per image. We \underline{underline} the best result for each model and display the top result for each metric in \textbf{bold}.
}
\label{fig:viz_results_5cap}
\end{table*}

\section{Human Evaluation}
\label{app:human_eval}

To conduct the human evaluation of the model generated responses we created a survey and hired 54 annotators through the Prolific platform compensated with 18\$ per hour. We added a screening in the platform for a representative sample of countries and an even distribution of male and female participants. Each annotator evaluated 12 images and their captions and for each image, we assigned two annotators and averaged their scores. We provided the following instructions to the annotators for evaluating the captions:

\begin{tcolorbox}[colback=white,colframe=blue,sharp corners]

This study involves evaluating captions. To guide your ratings,
imagine that you are describing the image to a visually impaired friend. Then consider: 

How well does the caption describe the image to this friend? Does it take into account cultural considerations? You will be given two sets of captions describing an image.

\begin{enumerate}
    \item Specify which caption you prefer for the given image (1, 2 or both).
    \item Determine if each caption is accurate and relevant to the given image.
\end{enumerate}

As a general guidance you should consider a caption as bad when it has one or more of the following
issues: 

a) Caption misses the main topic of the image.
b) Caption has major grammatical errors (such as being
incomplete, words in the wrong order, etc). Please ignore
the capitalization of words and punctuation. 
c) Caption includes hallucinations and mentions objects,
activities, or relationships that are definitely not in the
image. 
d) Caption is not as informative.
e) Caption does not reflect the cultural information depicted in the image.

\end{tcolorbox}

\section{Error Analysis II}
\label{app:errors}

\begin{figure*}
    \centering
    \includegraphics[width=\linewidth]{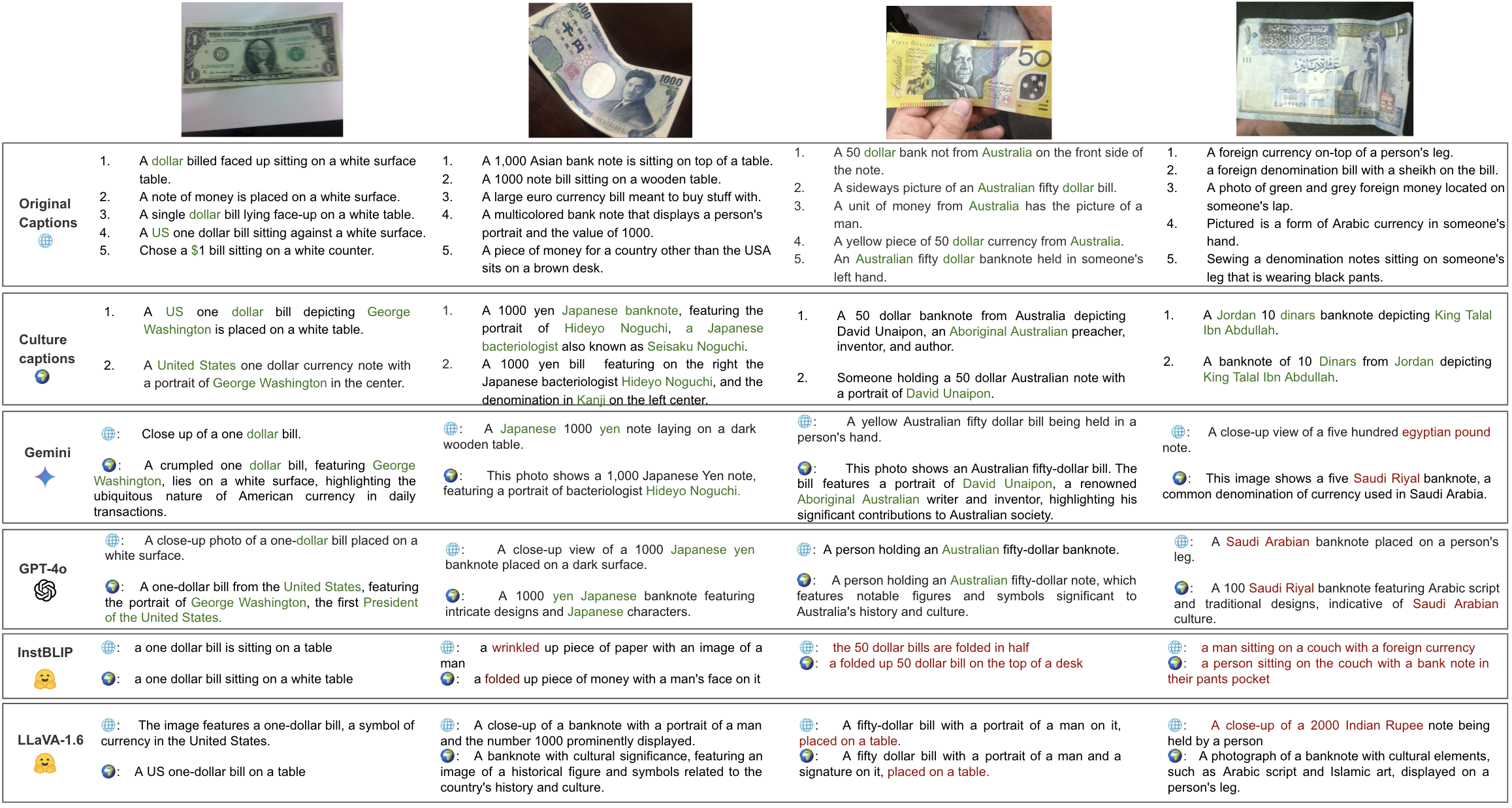}
    \caption{Examples from images related to currency comparing original (\emojidefault) with culture-specific (\emojiculture) annotations and generated captions from Gemini-Pro and GPT-4o with default (\emojidefault) and culture-specific (\emojiculture) prompting.}
    \label{fig:vizexamples}
\end{figure*}

We provide further examples from currency-related images in Figure \ref{fig:vizexamples}. We can see that for countries such as US, or Australia, the original VizWiz captions provide culture-specific information, but this is not the case for Japanese or Arabic currencies. Moreover, the models seem robust in western and Asian currencies, but not with all the Arabic ones. The example provided in Figure \ref{fig:vizexamples} shows how the models confuse a Jordan currency with Egyptian or Saudi Arabian currencies and how the smaller open-source models are more prone to hallucinations.

\section{Case Study}

We illustrate the value of cultural and inclusive VL models via a case study on evaluating GPT-4V as a visual assistant integrated into the ‘Be My Eyes’ platform. In this case study, we took a random sample of 20 images from the MaRVL dataset \citep{liu-etal-2021-visually}. Here we provide a selection of images we tried in our case study. Each figure includes the target culture behind each image and the GPT-4 Vision output after loading the image in the Be My Eyes application.

\begin{figure}[ht]
    \centering
    \includegraphics[width=\columnwidth]{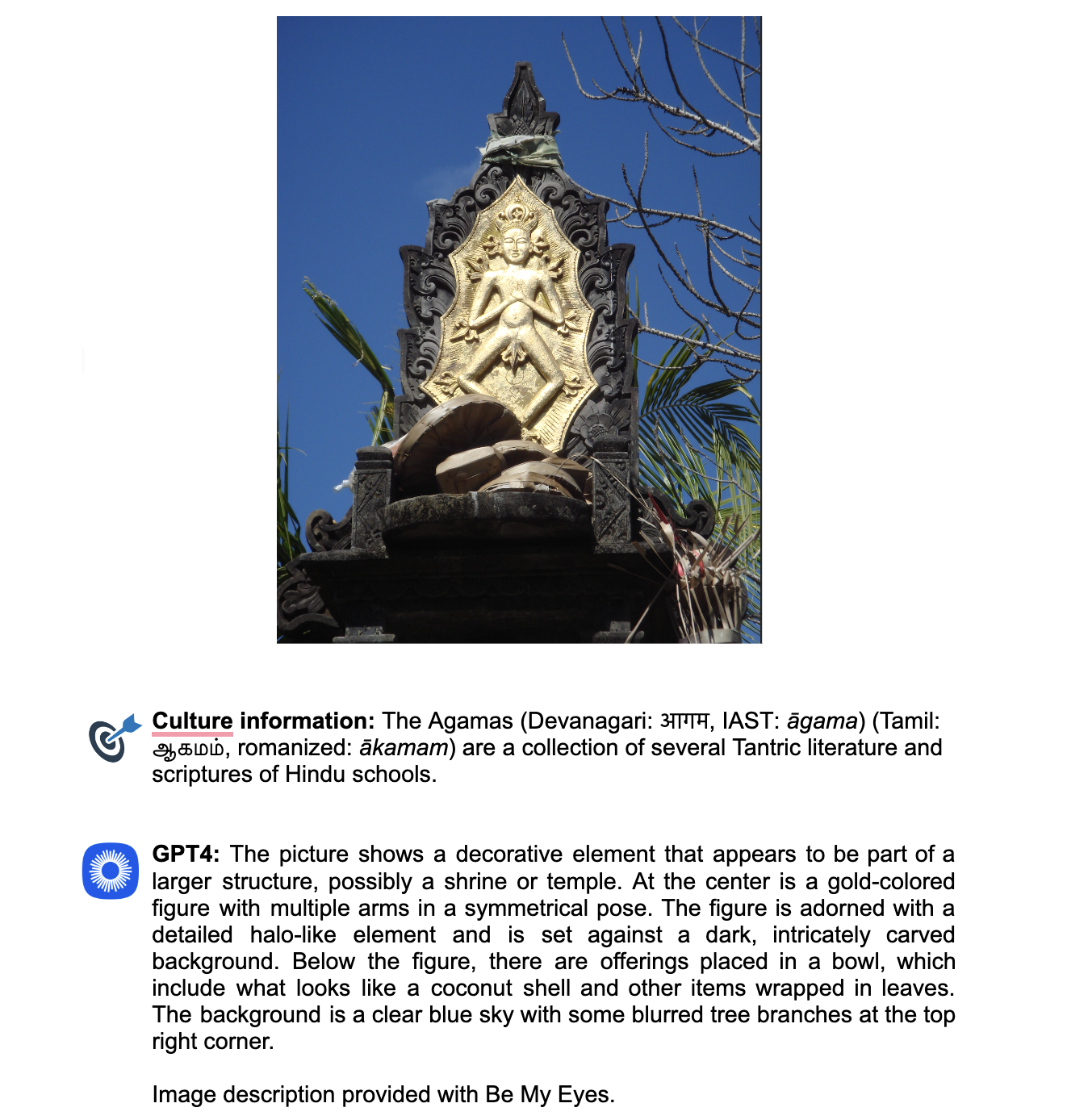}
    \caption{A picture extracted from MaRVL depicting a statue of Agamas and the GPT-4V image description provided in BeMyEyes.}
    \label{fig:mrvl1}
\end{figure}

\begin{figure}[ht]
    \centering
    \includegraphics[width=\columnwidth]{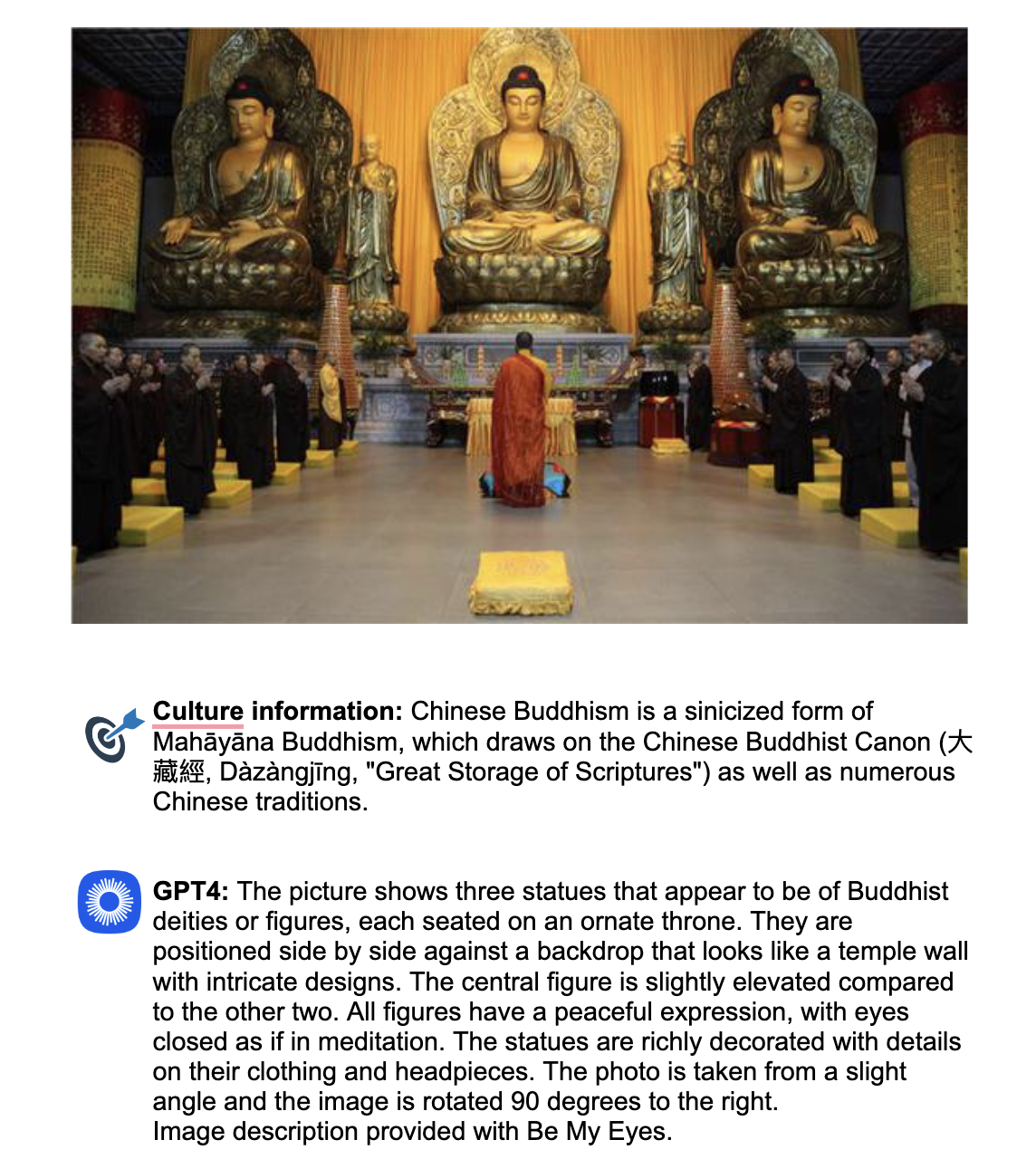}
    \caption{A picture extracted from MaRVL depicting Buddhist statues and the GPT-4V image description provided in BeMyEyes.}
    \label{fig:enter-mrvl2}
\end{figure}

\begin{figure}[ht]
    \centering
    \includegraphics[width=\columnwidth]{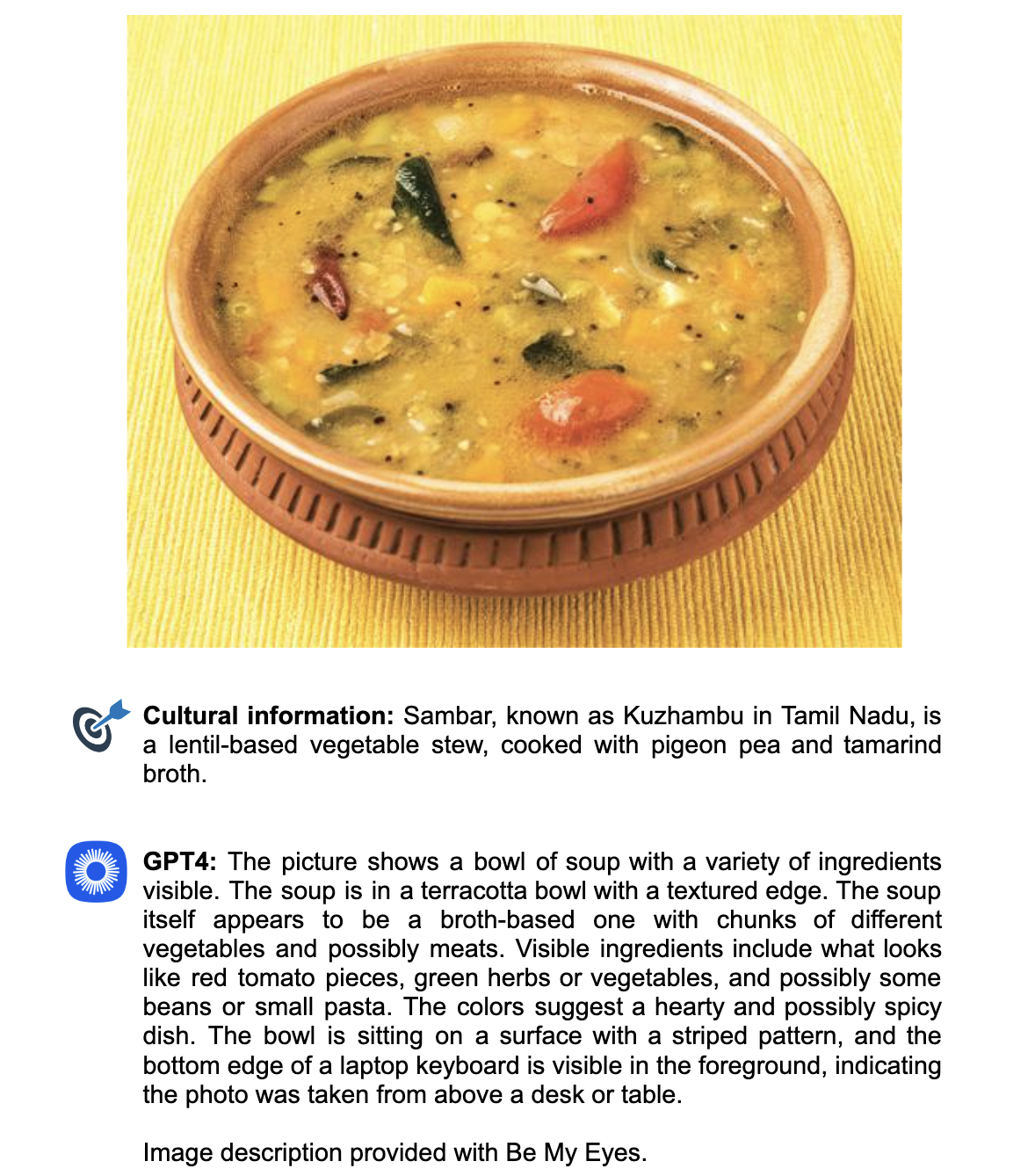}
    \caption{A picture extracted from MaRVL depicting sambar, a traditional Tamil dish, and the GPT-4V image description provided in BeMyEyes.}
    \label{fig:mrvl4}
\end{figure}

\begin{figure}[ht]
    \centering
    \includegraphics[width=\columnwidth]{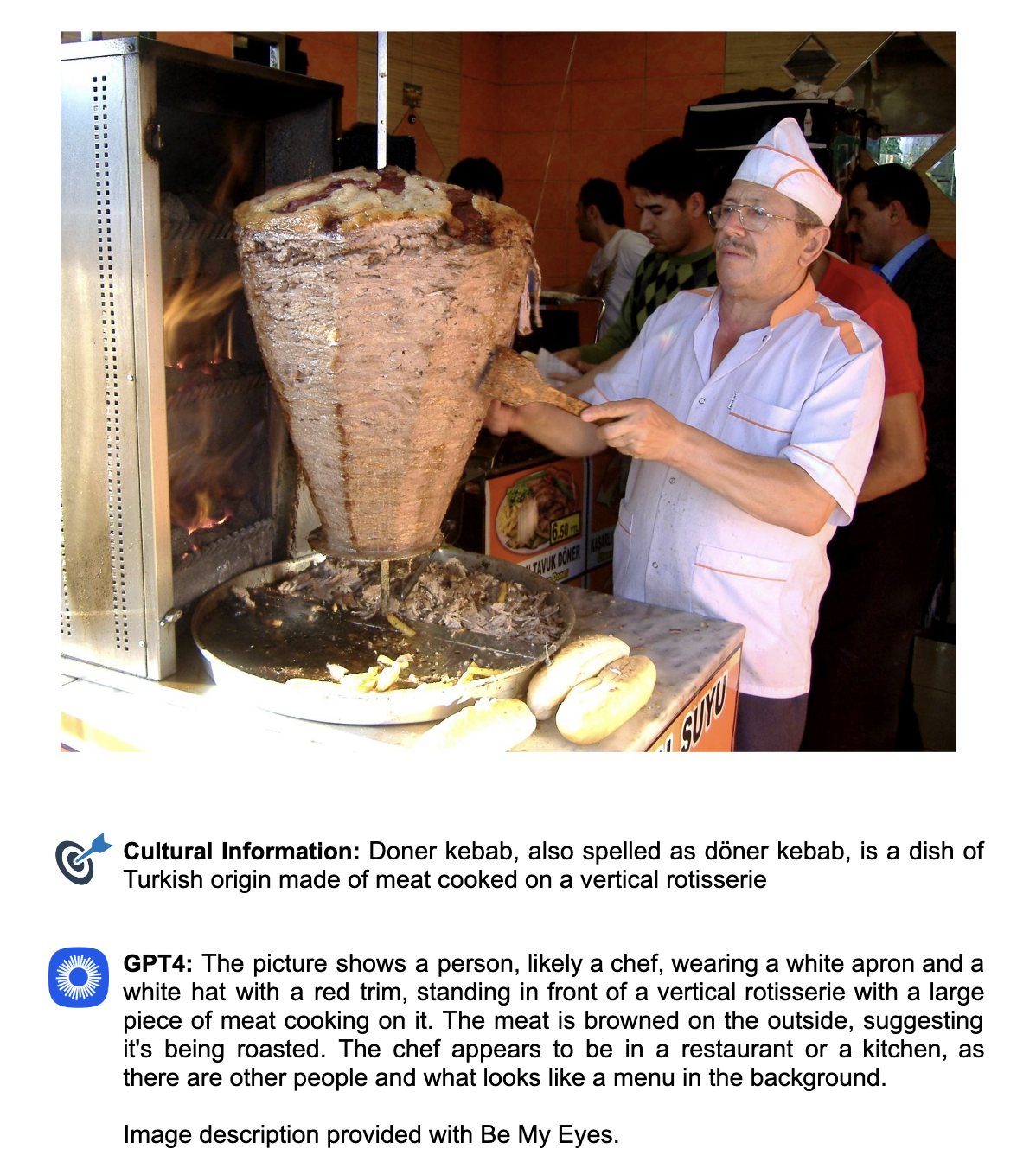}
    \caption{A picture extracted from MaRVL depicting a d\"{o}ner, a traditional Turkish dish, and the GPT-4V image description provided in BeMyEyes.}
    \label{fig:mrvl5}
\end{figure}

\end{document}